# From Scientific Controversies to Structured Understandings: Reflective Analysis in Artificial Intelligence, Cognitive Science, and Computing

**Paul S. Rosenbloom***

**Abstract**

Scientific controversies inherently reflect dissension but they also provide an opportunity to develop structured understandings that yield insights into both themselves and the disciplines within which they occur. This article builds on earlier work on three approaches to developing such understandings, plus a proposed unification of them, across the disciplines of artificial intelligence, cognitive science, and computing. The results here deepen the analyses of these approaches, including identifying their core operations and requirements for applicability, and a factoring of them that clarifies their structure while surfacing two additional cores; diagrammatically rethink their unification, as reflective analysis, in a manner that flexibly combines these five cores while providing an opening for additional ones; and contribute new applications of both individual cores and their unification.

## 1. Introduction

Scientific disciplines, unless in some idyllic fully agreed upon state, can be rife with controversies. At their worst, such controversies can threaten to tear apart the disciplines involved. The symbolic versus neural network controversy in artificial intelligence (AI) and cognitive science is one such example (Rumelhart *et al*, 1986b; Fodor & Pylyshyn, 1988). Both fields originated with a focus on symbolic processing as the basis for intelligence (Newell & Simon, 1976; Newell, 1980). With the neural network resurgence, and now dominance, one possible path would have been for new disciplines to have sprung up that are distinct from the existing ones. However, the two fields eventually became more accepting of other approaches, so what actually ended up occurring were within-discipline paradigm shifts (Kuhn, 1962).

Other types of controversies, such as about where to draw the boundaries around fields when it comes to hiring decisions, can lead to significant contention within an academic department. Smaller scale controversies, such as over the definition of intelligence (Legg, 2023), don't necessarily lead to such levels of contention, but can still result in serious confusion.

This work explores ways of analyzing scientific controversies to yield structures that aid in understanding not only the controversies themselves but also the disciplines within which they occur and the possible paths towards their resolution. Four types of

*Author Info: Professor Emeritus of Computer Science, University of Southern California, Los Angeles-USA.

controversies are considered here, concerning: (1) the meaning of specific concepts; (2) the types of elements used in theories, models, and systems; (3) the evaluation of theories, models, and systems; and (4) the standing of interdisciplinarity within scientific disciplines. The focus of this article is limited to controversies in AI, cognitive science, and computing, but by identifying the underlying requirements for these analyses the intent is to provide guidance as to where they might apply elsewhere.

Five general types of structures are leveraged here – sets, hierarchies, necessary features, maps, and languages – along with occasionally their combination. Sets provide the least structured approach, simply yielding extensional definitions of concepts in terms of the members of the sets. Hierarchies add part-of relationships among members of sets. Necessary features yield aspects of intensional definitions by focusing on what is in common among members of sets. Maps situate members of sets along one or more dimensions, highlight similarities and differences among them, and yield distance measures between them. They can thus reveal much about the nature of whatever is being mapped. Languages are more flexible than maps and as a result are capable of capturing much richer types of structures, albeit in a less orderly manner. The latter three approaches in particular can reveal much about the nature of the controversies, with the latter two approaches further capable of yielding insights into the structure of the disciplines within which the controversies occur.

This article builds on earlier attempts to understand the disciplines of AI, cognitive science, and computing by a professional computer/AI and cognitive scientist, not a professional philosopher, with initial results appearing in two books: *On Computing: The Fourth Great Scientific Domain* (Rosenbloom, 2012a) and *On Minds: Reflections of a Cognitive Architect* (Rosenbloom, 2026). The reflections in those books were based on three forms of analysis developed to aid in this understanding – *essential analysis* for concepts (Controversy Type 1), *dichotomic analysis* for elements (Controversy Type 2), and *relational analysis* for interdisciplinarity (Controversy Type 4) – and which were ultimately unified into a single *reflective method*. Evaluation (Controversy Type 3) was embedded within both essential and dichotomic analysis rather than being distinguished as a distinct form of analysis on its own.

The main contributions of this article are: the understanding that all of these forms of analysis fundamentally concern leveraging controversies to yield structured understanding; a deepening of the analyses of essential, dichotomic, and relational analysis, including identifying a more minimal *core* for each along with the requirements for its core to be applicable (Sections 2, 3 and 5); the clarification of the nature of the definitions that essential analysis provides for concepts and the different approaches that can be taken to developing these definitions (Section 2); the identification and analysis of *evaluative analysis* (Controversy Type 3) as a distinct form of analysis whose core may be combined with the others as appropriate (Section 4);[1]

[1] A fifth core for *cluster analysis* is identified in Section 6, but without much elaboration given its rather conventional nature.

two novel, albeit small-scale, applications of relational analysis (Section 5); a diagrammatic rethinking of the reflective method, as *reflective analysis*, that flexibly combines the existing cores while leaving room for additional ones that might later be identified (Section 6); and a new application of reflective analysis that further demonstrates its flexibility in structuring controversies (Section 7).

**2. Essential Analysis**

The core of essential analysis is a form of concept definition that combines an *essence* that is a consensus over necessity with a *definitional space* that captures the range of possibilities over which there is not (yet) a consensus (if in fact there are aspects that still lack consensus). In form, the approach is most similar to the neoclassical approach to concept definition (Leben, 2015), which combines a *skeleton* based on necessary features with *meat* of some form that fleshes out the skeleton. Here, the essence serves as the skeleton with the definitional space providing the meat. However, it is important to note that what is desired here is not a theory of human concepts but a useful means of defining controversial scientific concepts. Thus, the focus on consensus in determining the essence and on scoping out the remaining possibilities.

This notion of a consensus over necessity has been seen previously in AI and cognitive science in the development of the Common Model of Cognition (Laird *et al*., 2017), where it is used to characterize what must be in an architecture for humanlike minds – that is, the fixed structures and processes than enable human minds plus any other minds that can be modeled in a similar manner at an appropriate level of abstraction – based on a consensus across researchers, architectures, and data. The same general notion is applied here to the question of how to define controversial concepts, where the consensus is over alternative definitions of the concept. In particular, the idea is to identify an essence for the concept that is uncontroversial in its necessity while still capturing what particularly matters.

Achieving consensus on the Common Model involved not only consideration of existing architectures and results but also significant consultations with the community working on cognitive architectures. So far in this investigation, a short cut has been taken of considering existing definitions, when such exist, in combination with the author's judgement. More formal consultative processes can certainly be considered, and would be more appropriate in the long term, but have been beyond what have been attempted so far.

Examples of essences identified within the scope of this investigation have included *computing* as the transformation of information; *architecture* – whether of a building, a computer, or a mind – as fixity, or a bit more precisely as the fixed structures and processes that support variable content of the appropriate types; computational *symbols* as atoms or placeholders; *intelligence* as adaptivity; and *theory* as aboutness. Because these essences are intended to be consensuses over necessity, they do not in general capture the full richness traditionally associated with such concepts, but in exchange they cut through the controversies to capture some substantial commonality while delegating the more controversial remaining richness to their definitional spaces.

The definitional space is then typically a map of the broader range of possibilities not explicitly accounted for by the essence. Many of its dimensions may be based on the presence or absence of controversial aspects of existing definitions. For example, for symbols these aspects include composition (the ability to combine symbols into expressions), designation (the ability of expressions to indicate other entities, whether internal or external), interpretation (the ability of expressions to designate internal procedures for execution), and hybrid (the ability of expressions to have associated quantitative metadata; that is, numbers about them).

For symbols, the core of the traditional controversy turns out not to be whether such aspects are required – at least for the first three – but whether symbols with these aspects are necessary for minds, as claimed by the *Physical Symbol Systems Hypothesis* (Newell & Simon, 1976; Newell, 1980). Neural network approaches instead assume the necessity of metadata and deprecate the other three aspects. The new definition addresses this controversy by enabling the articulation of a *Hybrid Symbol Systems Hypothesis* as a proposed replacement for the Physical Symbol Systems Hypothesis (Rosenbloom, 2023), with the new definitional space for symbols now broad enough to support both traditional symbol systems and neural networks.

The definitional space for symbols also includes probabilistic systems (Pearl, 1988; Koller & Friedman, 2009) and a wide variety of hybrids, whether coarse-grained ones that simply combine conventional symbolic systems with neural networks or fine-grained ones that involve selective use of the optional aspects where appropriate. In the process it enables identifying which aspects of neural networks actually depend on classical symbolic aspects – for example, composition into vectors, designation by input nodes, and interpretation by hidden and output nodes – without assuming that all aspects of the system must be fully symbolic in the traditional sense. The overall controversy is thus reformulated somewhat less controversially in terms of which aspects of traditional symbols and neural networks are required where in yielding intelligence.

### 2.1. Approaches to Developing Essential Concept Definitions

Much of the original work on essential analysis assumed not only a specific concept representation but also a specific approach to developing concept definitions within that representation. In particular, the approach focused on concepts that had a range of existing definitions that were at least partially incompatible but that still shared some essence that could be extracted. That there was a controversy was obvious in such cases; and the definitions of the concept, rather than instances of it, served as the examples[2] from which the essence and definitional space would be derived. The derivations of the essence of *theory* as aboutness and of *intelligence* as adaptivity, both of which have many extant definitions, are prime examples of such an approach.

[2] In Rosenbloom (2026) examples were referred to as exemplars. That usage, which wasn't quite right, has been replaced here throughout.

Not all scientific concepts for which useful essential definitions can be derived precisely fit this mold, however. Five additional alternatives that were implicit in the earlier work are made explicit here: (1) where there are existing definitions, as in the standard approach, but there is also an unacknowledged lack of agreement as to what is even being defined (*intelligence*);[3] (2) where an implicit concept already exists but with no previous attempts at an explicit definition of it (*great scientific domains*); and (3-5) where there may not initially even be a notion that there are concepts to be defined (*core dichotomies*, *map locations*, and *linguistic expressions*). Section 2.1.1 covers alternative 1, Section 2.1.2 covers alternatives 2 and 3, and Section 2.1.3 covers alternatives 4 and 5.

### *2.1.1. Intelligence*

Intelligence was mentioned earlier as having an essence of adaptivity, but that glosses over an even deeper controversy in defining it – that the various definitions are not all actually about the same thing. Rosenbloom (2026) identified three distinct kinds of intelligence that existing definitions were focused on and provided a different essence for each: (1) *immediate*; (2) *cumulative*; and (3) f*ull-spectrum*. Immediate intelligence concerns only the here and now, with an essence of rationality. Cumulative intelligence focuses on learning and is effectively immediate intelligence plus lifelong improvement. Full-spectrum intelligence extensively itemizes the various capabilities necessary for intelligence, with a proposed essence of cumulative intelligence plus situatedness, (self-)awareness, projection of future possibilities, generality across problems, feelings, purposefulness, language, collaboration, socialization, and education. The notion of adaptivity was then tentatively proposed as the overall essence, based on analyzing the extent to which these three separate essences could be reduced to some very general notion of it.

The three levels of definitions also complicate determining what the definitional space for intelligence is. Each of the two higher levels encompasses the lower levels but also includes aspects that were considered optional in them. Thus, each could be considered as yielding a definitional space for the levels lower than it, with full-spectrum intelligence possibly having a definitional space that includes additional aspects not explicit in its list.

There is also a separate notion of an *intelligence space* that attempts to cover the range of possible intelligent entities, or at least much of it, via approximations to the three essences. While acknowledging the necessity of the essential definitions it assumes that perfect realization of their essences is not necessary, opening up room for the wide variety of both natural and artificial entities that approximate these ideals in one way or another. This is thus a space of variations within what is considered necessary rather than without it.

[3] Consciousness is another concept with a similar, if not even worse, problem but not yet one that has been tackled directly in this fashion.

*2.1.2. Great Scientific Domains and Core Dichotomies*

The concept of a great scientific domain is implicit in the everyday notion that there exists the physical, life, and social sciences. However, in pondering where computing fits in the overall space of intellectual endeavors, it became necessary to ask whether there was a definition implicitly in common among these three existing domains that would also apply to computing. The essence here is one of the more involved ones: the understanding and shaping of the interactions among a coherent, distinctive and extensive body of structures and processes. In this form, science spans many intellectual endeavors not normally considered part of it, including philosophy and mathematics. *Metascience* – that is, the science of science (e.g., Metascience, 2026) – is thus also broadened as a side effect, to include philosophy of science, with the forms of analysis studied in this article thus fitting within both of these categories; and these studies themselves serving as forms of science of metascience, or meta$^2$science.

This essence of science actually gets close to a classical definition in form, but it does omit some of what were considered to be the more controversial aspects of many traditional definitions, such as the method(s) used. A form of methodological pluralism then became part of the definitional space, although with the addition that, for any particular question, there should be a preference for those applicable methods that are strongest in terms of the veracity they can achieve. In consequence, the methods used form an important component of the definitional space of a great scientific domain.

The other nonstandard part of this essence is that it includes not only understanding but also shaping, which corresponds to such activities as engineering for the physical sciences, medicine for the life sciences, and business and education for the social sciences. The justification for this was the assumption that just as understanding and shaping are closely intertwined in computing, where in fact much of what is to be understood is what we as humans have shaped – that is, what is "artificial" (Simon, 1969) rather than "natural" – this would be more and more true of the other domains as they become increasingly able to shape their own phenomena at their most elementary levels.

A dichotomy – such as procedural versus declarative memory (e.g., Squire, 2004) or heteroassociative versus autoassociative memories (e.g., Bates, 2020) – is a contrast between two options, and thus may naturally indicate a controversy all by itself. This use of dichotomies is front and center in dichotomic analysis (Section 3), where each dichotomy is to independently bifurcate the space of elements out of which theories, models, or systems are built so as to jointly yield a multidimensional map of this space. However, there can be too many dichotomies overall to make this practical directly; in the work on dichotomic analysis, eighty-eight dichotomies concerned with technologies from AI, cognitive science, and computing were included, with still more yet to be considered. Moreover, many of these dichotomies turned out to be related in one way or another.

So, dichotomic analysis presumes that a relatively small number of sets of related dichotomies can be identified, with each such set amounting to an extensional definition of a concept that can be reduced to a single *core dichotomy* that is its essence. Although this may sound like prototype theory (Rosch, 1973), what is sought here are core dichotomies that capture the necessary aspects of set membership rather than prototypical aspects, and there is no notion of distance between a core dichotomy and the other members of its set. Instead, a core dichotomy is a neoclassical definition of its concept. The controversy then concerns what dichotomy can best act as this core.

Preexisting concepts with a range of known definitions did not exist for these sets, but essences, which are themselves dichotomies, are still required over the confusion of dichotomies within each set. The definitional space for such a set can then be specified extensionally by the original set itself, or if more fundamental dimensions of variation over the instances can be identified then a more intensional definition of the space may be possible.

In the work on dichotomic analysis three sets of related dichotomies were identified, and an essence was determined for each such set. The first set, comprising thirty-one dichotomies, yielded asymmetric versus symmetric, or *(a)symmetric*, as its essence. The two example dichotomies seen earlier are both members of this set. The second set, comprising thirty-two dichotomies, yielded uncontrolled versus controlled, or *(un)controlled*, as its essence. Examples of it include System 1 versus System 2 thinking (Kahneman, 2011) and backpropagation (Rumelhart *et al*, 1986a) versus reinforcement learning (Sutton & Barto, 2018). The third set, comprising twenty-five dichotomies, yielded metadata versus data, or *(meta)data*, as its essence. Examples of it include utilities versus goals and neural networks versus rules.

Both the standard approach to developing essential definitions, of analyzing existing definitions, and the two alternatives in this section conceptually involve a form of induction from examples in searching for necessary conditions, although the examples in the standard approach are existing definitions of a concept whereas here the examples are instances of a concept with no previous definition. This illustrates two different, yet informal, inductive approaches to developing essential definitions. The definitional spaces in both approaches are based on variations in the examples upon which the definitions are based that go beyond the essence. This space would be defined extensionally if it just amounts to the examples themselves, or it can be defined intensionally if more fundamental dimensions of variation are derivable from the examples.

### *2.1.3. Map Locations and Linguistic Expressions*

Defining concepts in terms of either locations in maps or specific linguistic expressions involves more deductive approaches. Given dimensions of a space – whether based on dichotomies (dichotomic analysis) or scientific domains and relations among them (relational analysis) – a combination of locations along each dimension effectively defines a concept, such as asymmetric, noncontrolled, metadata technologies (e.g., feedforward neural networks) or the implementation of intelligence – part of the social domain (S) – by computing (C) (e.g., AI). Given a language, such as the metascience expression language (Section 5), each expression in the language effectively defines a concept (e.g., S/C, for AI's implementation of the social domain by computing). These concepts can then be populated by examples that fit their description, raising the possibility of either extensional or intensional specifications of their definitional spaces.

### 2.2. Applicability Criteria for Essential Analysis

For essential analysis to be applicable, there must be necessary conditions at the heart of the concept that can be extracted from the initial information that is available about it. As has been seen, this initial information can, at least, take the form of prior definitions, instances, map locations, or linguistic expressions. There is also an expectation that a consensus be developed around the essence, although that has not been a focus here. With respect to the definitional space, an extensional definition requires examples whose variations go beyond the necessary conditions. An intensional definition requires that these examples embody dimensions of variation that go beyond the necessary conditions.

Within these constraints, excepting the need for a consensus, essential analysis has so far been demonstrated to be applicable to both the definition of new terms and the redefinition of existing ones, with these terms ranging from relatively narrow technical concepts, such as symbols and architectures; to broader concepts, such as intelligence and theories; to dichotomies and categories of technologies; to whole theories, such as the Common Model of Cognition; to the nature of entire disciplines, such as computing, and its subdisciplines; and to even broader segments of the world of intellectual investigation, such as science, engineering, and great scientific domains.

## 3. Dichotomic Analysis

The core of dichotomic analysis is leveraging one or more dichotomies to define a space in which the elements out of which theories, models, or systems within a discipline are composed. The result is a map of the discipline within which these elements can be situated, and which can potentially help in understanding the discipline as a whole.[4] For AI and cognitive science, the elements of interest have been individual technologies, such as feedforward neural networks and rules, which can act as elements both of theories/models of human cognition and of AI systems.

---

[4] Although the core of dichotomic analysis defines the structure of this map, it doesn't itself populate it with elements. How this population occurs will become clearer when reflective analysis is introduced in Section 6. Still, partial populations are assumed as necessary for illustration in the remainder of this section.

Section 2 discussed how three sets of dichotomies across AI and cognitive science were each reduced to a core dichotomy – (a)symmetric, (un)controlled, or (meta)data – that bifurcates the disciplines. Each core dichotomy thus effectively yields a 1D map with two locations. Crossing two core dichotomies – (a)symmetric × (un)controlled, (a)symmetric × (meta)data, or (un)controlled × (meta)data – yields a 2D map with four locations, and crossing all three dichotomies – (a)symmetric × (un)controlled × (meta)data – yields a 3D map with eight locations.

The "origin" corner of such a 3D map, based on asymmetric, uncontrolled, metadata, contains for example heteroassociative – that is, feedforward – neural networks, whereas the opposite corner, based on symmetric, controlled, data, contains for example planning. Intermediate locations contain less "extreme" technologies, such as Soar's parallel rule system (Laird *et al.*, 1987; Laird, 2012) at asymmetric, uncontrolled, data, and autoassociative neural networks such as Boltzmann machines (Ackley *et al.*, 1985) at symmetric, uncontrolled, metadata.

Such maps directly support inferences concerning similarities, differences, and distances – measured as the rectilinear, or Manhattan, distance – among technologies. Feedforward neural networks and planning are, for example, opposites that differ along all three dimensions and thus are at a distance of three. The same is also true for some less obvious pairs of technologies, such as Boltzmann machines (symmetric, uncontrolled, metadata) versus decision trees (asymmetric, controlled, data) and monotonic logic (symmetric, uncontrolled, data) versus nonmonotonic numeric functions (asymmetric, controlled, metadata). Other technologies that have generally been considered to be quite different, such as feedforward neural networks (asymmetric, uncontrolled, metadata) and Soar's rules (asymmetric, uncontrolled, data) turn out to be only at a distance of one, with such rules unexpectedly being closer with respect to such a map to neural networks than they are to planning (symmetric, controlled, data).

Experience to date has shown that most technologies in AI and cognitive science map directly to individual locations in the 3D map, although some complex ones must be decomposed into smaller parts that can fit at single locations, and others are sufficiently broad in applicability so as to span multiple locations without being decomposed in such a coarse-grained manner. One prototypical complex example is analogy (e.g., Falkenhainer *et al.*, 1989), which involves symmetric processing to retrieve possible analogues of a target followed by asymmetric processing to map results from the analogue to the target. One broad example is Sigma's generalized factor graphs (Rosenbloom *et al.*, 2016), which span the entire uncontrolled region of the map. Within a single unified algorithm, microvariations – such as the directionality of message passing and the relative importance of data versus metadata – determine in which regions of the map portions of the extended factor graph operate.

Beyond individual technologies are entire systems that combine multiple technologies – such as cognitive architectures (Langley *et al.*, 2009; Kotseruba & Tsotsos, 2020) and hybrid neural systems (e.g., Silver *et al.*, 2018; Hitzler & Sarker, 2022) – and which map to a combination of regions, much like complex or broad individual technologies but even more so. Consider, for example, a deeper dive into Rosenbloom's (2026) comparison of the Common Model of Cognition (Laird *et al*., 2017) and AlphaZero (Silver *et al*., 2018). On the surface these are two quite different systems. The Common Model, as already described, is an abstract consensus concerning the structures and processes required in a humanlike cognitive architecture. It covers the union of the asymmetric and uncontrolled halves of the map. In contrast, AlphaZero is an AI program that uses neural networks, reinforcement learning, and Monte Carlo tree search to learn to play board games such as chess and go. It covers the controlled half of the map plus the asymmetric, uncontrolled, metadata location.

One way to characterize what these two systems have in common is that they both cover the asymmetric half of the map, incorporating some form of procedural memory that leverages to some extent both data[5] and metadata (asymmetric, uncontrolled) plus sequential action with reinforcement learning (asymmetric, controlled). Such capabilities are essential to any system that is to act in the world. However, beneath this commonality the systems do differ in some details with respect to these capabilities: in procedural memory AlphaZero places much more emphasis on metadata, via feedforward neural networks, whereas the Common Model places much more emphasis on data, via rules; and the Common Model incorporates much more flexibility in terms of both the kinds of actions that can be performed and the kinds of information that can be used for selecting among them, with reinforcement learning yielding only one form of the latter.

Beyond this commonality and second-order differences, each system includes a symmetric component, although AlphaZero's is controlled (Monte Carlo tree search) whereas the Common Model's is uncontrolled (declarative memory). For its games, AlphaZero needs search, a form of reflection that enables it to engage in extensive reasoning in support of action selection, although its Monte Carlo tree search is far from a general reflective capability. At the same time, because of its limitation to closed-world game playing, it has little need for the kinds of general knowledge about the world that would be stored in a declarative memory. The Common Model does require a full reflective capability, and proposals exist for how this could work (e.g., Laird *et al*., 2026), but no consensus has yet been reached on it, so it is presently absent from the model. The Common Model also requires a full complement of long-term memories, and because a consensus does exist on declarative memory, it has been included.

### 3.1. Applicability Criteria for Dichotomic Analysis

[5] This assumes that neural networks do include some minimal amount data – that is, their nodes – as with the essential analysis of symbols in Rosenbloom (2023), so that they, again minimally, cover asymmetric, uncontrolled, data.

For dichotomic analysis to be applied to a discipline it must be possible to identify a set of core dichotomies over its elements.[6] For AI and cognitive science (and, to a lesser extent, computing) this was accomplished by amassing eighty-eight initial dichotomies over technologies that clustered into three sets and using essential analysis to derive a core dichotomy for each of these sets.

A second criterion that matters is the degree to which the core dichotomies together span the discipline of interest. For AI and cognitive science this was plausible up front due to how the three core dichotomies captured the essences of so many individual dichotomies – thirty-one, thirty-two, and twenty-five respectively – from across the two fields (and occasionally even computing). However, a more rigorous test would be to identify all of the relevant technologies within the disciplines and determine whether they fit somewhere in the 3D map. While the list of technologies considered to date is not exhaustive, none has so far been identified that could not be handled in this manner. Yet, even if a map in general does not cover all elements of a discipline, covering a critical mass may still be sufficient for utility. So, the true requirement is that the core dichotomies cover enough of the discipline to be useful.

## 4. Evaluative Analysis

The core of evaluative analysis is comparative evaluation of the elements within a space via *a full array of relevant criteria*. Accuracy – with respect to phenomena for theories and models, and with respect to benchmarks for systems – is such an important criterion that it often is implicitly assumed to be the only one that matters. However, scope of applicability can challenge accuracy for supremacy in areas such as cognitive architectures, where scope is their reason for being and trying to achieve scope before refining the result for accuracy may be a better research strategy for combining the two criteria than optimizing accuracy from the beginning. Similarly, clarity may be critical when the goal is education rather than prediction.

There is a rich history of developing extensive lists of evaluation criteria for both cognitive architectures and the whole minds they are to support (e.g., Newell, 1980; Newell, 1990; Anderson & Lebiere, 2003; Sun, 2004; Langley *et al*., 2009; Varma, 2011; Goertzel, 2014; Kotseruba & Tsotsos, 2020; Lieto, 2021). Grounded in this earlier work, Rosenbloom (2026) developed five distinct but related lists of evaluation criteria, for: (1) theories; (2) models of general intelligence; (3) cognitive architectures; (4) the Sigma cognitive architecture; and (5) AI technologies and systems. The goal here is not to make sense of such a plethora of lists but to introduce evaluative analysis as a distinct form of analysis that is based on more than just evaluation of accuracy, while simultaneously rescuing it from its previous subservient status within essential analysis (lists 1-4) and dichotomic analysis (list 5).

---

[6] Of course, larger sets of alternatives that yield trichotomies and even higher order -otomies can also be considered, enabling finer grained maps; but only dichotomies have been found necessary so far.

Evaluative analysis in the context of essential analysis concerns relative criterial advantages within definitional spaces of concepts. As discussed in Section 2, a critical part of the standard approach to essential analysis is stripping out what might previously have been considered definitional aspects of a concept – such as accuracy for theories and composability for symbols – and converting them instead into optional aspects that become dimensions of the definitional space. Some of these aspects/dimensions may have preferential directions that can be leveraged as evaluation criteria, such as accuracy and lawfulness for theories, while others – such as composability and designation for symbols – may not. In addition to the evaluative dimensions that result from this process, other criteria may also be added post hoc. For theories, for example, this might include such criteria as comprehensiveness and surprisingness. Identifying a full list of such criteria enables understanding the interplay among the criteria and how desirability may vary across the definitional space.

Evaluative analysis arose less directly in the context of dichotomic analysis, while ultimately yielding more complex results. Instead of individual dimensions of the space directly yielding evaluation criteria, as in essential analysis, multiple evaluation criteria may be relevant to each dichotomy. A hierarchical list of evaluation criteria was developed – list 5 – which at the top level includes accuracy, efficiency, generality, simplicity, improvability, usability, and operability. Having multiple criteria can lead to tradeoffs along each dimension, where one criterion may be better met at one end while a second is better met at the other end, thus motivating attempts at spanning the associated dichotomies so as to achieve the best of both worlds. Prototypical examples include combinations of procedural and declarative memories (e.g., Squire, 2004; Laird *et al.*, 2017) for (a)symmetric, dual-process theories (e.g., Kahneman, 2011; Evans & Stanovich, 2013) for (un)controlled, and neuro-symbolic hybrids (e.g., Sun, 2016; Hitzler & Sarker, 2022) for (meta)data.

Even more complex tradeoffs appear when combinations of dimensions are considered together in multidimensional maps. The result is maps of criterial advantages rather than maps of technologies. In Rosenbloom (2026) these were referred to as a second type of map produced by dichotomic analysis. However, in distinguishing evaluative analysis from dichotomic analysis, while using the same overall map structure, it becomes clear that each form of analysis yields exactly one type of map.

The complexity of analyzing evaluative maps with many criteria led to the development of the *Criterial Advantage (CA)* language for expressing compactly the criterial advantages of regions of dichotomic maps. This is a compositional language that enables independently expressing the criterial advantages associated with individual dichotomies – that is, which criteria are differentially advantaged at each end – and then combining them as dichotomies are crossed in generating higher dimensional maps. In the process it enables comparing individual technologies at different locations in a map as well as systems that combine technologies spanning whole regions of a map.

As an example, let's consider further the comparison of the Common Model of Cognition and AlphaZero that was begun in Section 3. By combining the impacts of all three dimensions on the individual locations of its 3D map and then aggregating across its non-empty regions, the Common Model was summarized in the CA language as $[A(P^2, C^2, R), E(T(C_{(1^2, n)}, R^3), S_{(1, n^2)}), G(P_{(a^3, gs^2, gl^3)}, E), IL^3, U^2, O^2]$.

There is much to unpack here. First, there is a hierarchy of advantages, with each represented by its initial letter (unless that leads to ambiguity). So, for example, $A(P^2, C^2, R)$ represents Accuracy, with its three subcriteria of Precision, Correctness, and Robustness. Second, superscripts denote the number of "doses" of an advantage that result from multiple dimensions contributing to it. For any possible advantage there is thus between zero and three doses in a 3D map, with zero corresponding to a complete lack of that advantage. Precision and Correctness each have two doses here whereas Robustness only has one. Third, subscripts denote restrictions on the scope of an advantage. For example, for ETC – that is, Combinatoric Temporal Efficiency – a subscript of "1" indicates an advantage when a single solution is sought while a subscript of "n" indicates an advantage when multiple related solutions are sought. Thus, there are double doses here for a single solution and only one dose for multiple related solutions.

There is more detail that could be explored in this expression, but this is enough to enable a useful comparison of criterial advantages between the Common Model and AlphaZero, whose expression was $[A(P^2, C^3, R), E(T(C^2, R_{(1c^3, 1d^2, n^2)}), S_{(1, n^2)}), G(P_{(a^3, gs^3, gl^2)}, E), IL_{(a^3, g^2)}, U^2, O(Ef^2, R^2, Sa^3)]$. This comparison reveals that the Common Model does better on some special cases of real-time efficiency, one special case of problem generality, and one special case of learnability. In turn, AlphaZero does better on correctness, one special case of combinatoric temporal efficiency, one special case of problem generality, and safety. These differences in criterial advantage are directly due to the differences in coverage discussed in Section 3, with the Common Model's lack of a reflective capability having the single largest impact. Still, at a high level there is much more in common across these two rather disparate systems than there are differences.

### 4.1. Applicability Criteria for Evaluative Analysis

To what extent could evaluative analysis be applied in other disciplines? The simpler form of it that was used in the context of essential analysis is potentially applicable whenever dimensions of a definitional space have preferential directions. The analysis of theories itself should be applicable in whatever disciplines theories appear, although the importance of particular dimensions, and thus the tradeoffs across the dimensions, may vary by discipline. Beyond this, the general case, which subsumes both the simpler and more complex forms, can be seen to depend on having maps for which tradeoffs among one or more evaluation criteria can be identified along one or more dimensions.

## 5. Relational Analysis

The core of relational analysis is structuring the interplay among scientific disciplines in a manner that clarifies the interdisciplinary nature of their subdisciplines and the similarities and differences among them. This structuring is expressed in general via the *Metascience Expression (ME)* language. The language is built around abbreviations for the disciplines involved, which just amounts to their first letter when it is unambiguous. In Rosenbloom (2012a) these were primarily the four great scientific domains – the [P]hysical, [L]ife, [S]ocial, and [Co]mputing sciences – however, subdomains such as [M]athematics and [H]umanities (Rosenbloom, 2012b) can also be considered, as can a notion of any or all sciences (Δ). In general, a discipline at any level of granularity can conceivably be used.

The disciplines are related through a pair of relations: *implementation* (/) and *interaction* (↔,→ and ←). Implementation reflects how one discipline may implement another, as in how the physical sciences naturally implement the life sciences (L/P) and the life sciences naturally implement the social sciences (S/L). However, it can also represent such things as how the physical sciences implement computing (C/P) in computer hardware, and how computing implements the social sciences (S/C) in AI. Implementation thus operates across levels of science, and is essentially the dual of the traditional notion of *reduction* (van Riel & van Gulick, 2025). Interaction is instead a peerwise relation at a single level that can be bidirectional, as in human computer interaction (S↔C), or unidirectional, as in computer vision (C←Δ) or manipulation (C→Δ).

The full ME language can be reduced down to a 2D map if a single discipline is the focus and only its relationship with one other discipline at a time is considered. In such a case one dimension of the map spans the other disciplines to which the focal discipline relates, and the second dimension spans the types of relations between the disciplines. In such a map for computing, each location includes the (sub)disciplines within computing that are best characterized by the combination of the specified relationship and the other domain. A map of this sort was developed for *dyadic computing* – that is, computing in relation to one other domain – in Rosenbloom (2012a). Unanticipated there, however, was that it is also conceivable to develop a 3D map that covers all of *dyadic science*, where two dimensions would each span the great scientific domains, with the remaining dimension spanning relationships between pairs of these domains.

Extrapolating from dyadic science, it is possible to talk about *polyadic science*, which involves relationships more generally among multiple domains or disciplines.[7] This can involve single instances of multiple domains/disciplines, such as S/L/C for the computational (C) implementation of networks of neurons (L) that can yield intelligent behavior (S), or (S/C)→P for intelligent (S) computational (C) systems manipulating the physical (P) world. But it can also involve more than one instance of a single domain if it plays multiple roles, such as S/(C↔L/C) for hybrid models of intelligence (S) implemented (/) by interactions (↔) between symbolic (C) and neural (L/C) technologies. Polyadic science leverages the full capabilities of the ME language, enabling representing complex topics that are difficult to display in maps, such as (L/(S/C↔P))↔P for intelligent robots (L/(S/C↔P)) – that is, implementations (/) of (simulations of) life (L) by interactions among smart computational (S/C) and physical (P) components – interacting (↔) with the physical world (P).

[7] Based on Rosenbloom (2012a), the more limited notion of *polyadic computing* has recently been proposed as the basis for the field of *informatics* in Europe (Zatsman, 2023).

Two novel applications of relational analysis relevant to the work here are to: (1) structure the approaches that can be used to span dichotomies, such as those mentioned in Section 4, to yield hybrids of their two poles; and (2) provide some clarity as to the ways of drawing boundaries around disciplines that maintain rich relationships with other disciplines. The first of these was anticipated in Rosenbloom (2012a) but not fully developed there. If the two poles are represented abstractly as A and B, the general case can be denoted in ME simply as A+B, which states that both poles exist in some unspecified relationship. From there, two general approaches to hybridization have been explored: *addition*, which simply includes modules for both poles; and *reduction*, which finds a commonality between the two poles that enables them to be supported via a single module (Rosenbloom, 2011).

Addition can be denoted as (A↔B) if the two modules not only exist side-by-side but also interact, or as {A, B} if they instead simply form an unconnected set. As mentioned earlier, reduction is the dual of implementation, so it can be represented in one of three forms: A/B, B/A or (A↔B)/C). The first two forms simply reduce one pole to the other; or, equivalently, they implement one pole by the other, as for example when symbolic processing is implemented via neural networks. The third form arises when both poles are implemented by a common underlying functionality, such as how neural and symbolic computing both arise from Sigma's extended factor graphs (Rosenbloom *et al*, 2016).

The other novel application, while it doesn't resolve the question of where boundaries should be drawn around fields with rich interdisciplinary connections, it does at least systematize the options for doing so. There appear to be three general strategies for drawing such boundaries. The first strategy focuses only on a pure notion of what should be included. For computing, this would mean that the field can be represented simply by the monadic expression C, including such subdisciplines as analysis of algorithms and programming languages, but leaving out much of the richness of what would typically be found in a computer science department. Such a strategy thus explicitly relegates all interdisciplinary work to other unspecified – and possibly nonexistent – fields, with enhanced structural difficulty in building bridges among them.

The second strategy includes the pure notion but further includes any interdisciplinary work that is "in service" of the field. The logic behind this is that such additional forms of work are about the discipline rather than about some other discipline. For computing this would thus include (sub)disciplines of the form C←Δ and C/Δ, for arbitrary Δ, such as computer vision (C←P) and computer hardware (C/P). Relationships in the other direction instead may be dismissed as applications in service of some other discipline rather than as study of the discipline itself. I have heard computer scientists try to make this distinction, but it can lead to unexpected conclusions, such as that AI (S/C) and robotic manipulation (C→P) should not be considered as part of the field.

The third strategy includes all (sub)disciplines that have the field anywhere in their ME expressions, whether traditionally considered basic science or applications. This yields a full polyadic field, assuming monadic is included as the limiting case of polyadic, with the implication that disciplines will overlap whenever their (sub)disciplines share ME expressions. This can be thought of as drawing the boundaries around fields as broadly as possible, or as including the bridges among fields as parts of the fields themselves. It also harks back to the combination of understanding and shaping found in the definition of great scientific domains (Section 2.1.2).

This way of viewing the three strategies doesn't resolve controversies about boundary placement, but it does clarify them. The first is an isolationist strategy that attempts to retain a notion of purity about the nature of the discipline, whereas the third is open to all possibilities that include the core discipline. The second is a midpoint between the other two, embodying a "selfish" notion of just what matters to the core discipline. As someone who has worked across disciplines throughout my career, I am a strong proponent of the third approach, as a means of facilitating rather than impeding interdisciplinary work, but this is far from a universal perspective.

### 5.1. Applicability Criteria for Relational Analysis

Application of relational analysis to a discipline depends on it having significant relationships with other disciplines. But the disciplines involved can potentially range from great scientific domains down to arbitrary subdisciplines. So far, no further types of relationships beyond interaction and implementation have proven necessary; and how these two handle both relationships at a single level via interaction and relationships across levels via implementation would appear to make the need for others unlikely.[8] Still, should such prove necessary, extension of the ME language to incorporate them would always remain a possibility.

## 6. Reflective Analysis

The reflective method was developed as a unification over essential, dichotomic, and relational analysis in an attempt to yield a single broadly applicable analytical method for use in understanding scientific topics and structures composed from them. It is a relatively weak metascientific method, in terms of the veracity it can guarantee, intended for problems for which there are presently no stronger methods. As defined in Rosenbloom (2026) it consisted of a sequence of six steps:

(1) Identify *topics* of interest.

(2) Amass *examples* of these topics.

(3) Identify the *essences* of these topics.

(4) Identify *structures* within and/or across these topics.

(5) Assort *examples* of these topics across these structures.

(6) Derive *additional implications* from the previous steps.

---

[8] At an early point in the development of relational analysis a third type of relationship – *embedding* – was also included, but it was dropped when it became clear that it could be reduced to a particular combination of interaction and implementation. For example, a prosthetic – that is, a physical device embedded within a living organism – can be represented as L/(L↔P) to reflect that a form of life (L) is implemented by interactions between the life (L) and physical (P) domains.

It was mentioned, however, that these steps need not proceed exactly in this order, nor must they be unitized. Some of the activity at the heart of Step 2, of amassing examples, could for example occur before or during Step 1 or during Step 5 – once the structures are identified in Step 4, more examples could be amassed while assorting them across these structures. Likewise, some of the additional implications at the heart of Step 6 could occur earlier.

The complexity of ordering the steps in the reflective method, along with the broader question as to whether it is truly a method in the philosophical sense rather than just in the looser sense common in computing, have led to the identification of the cores found in the previous sections and to the diagrammatic rethinking of the reflective method, as *reflective analysis*, that is shown in Figure 1. Here, there are no steps, just processes, with information flow – as indicated by the arrows – among them. The six steps of the reflective method are reduced here to four processes, with Steps 1 and 3 mapped to

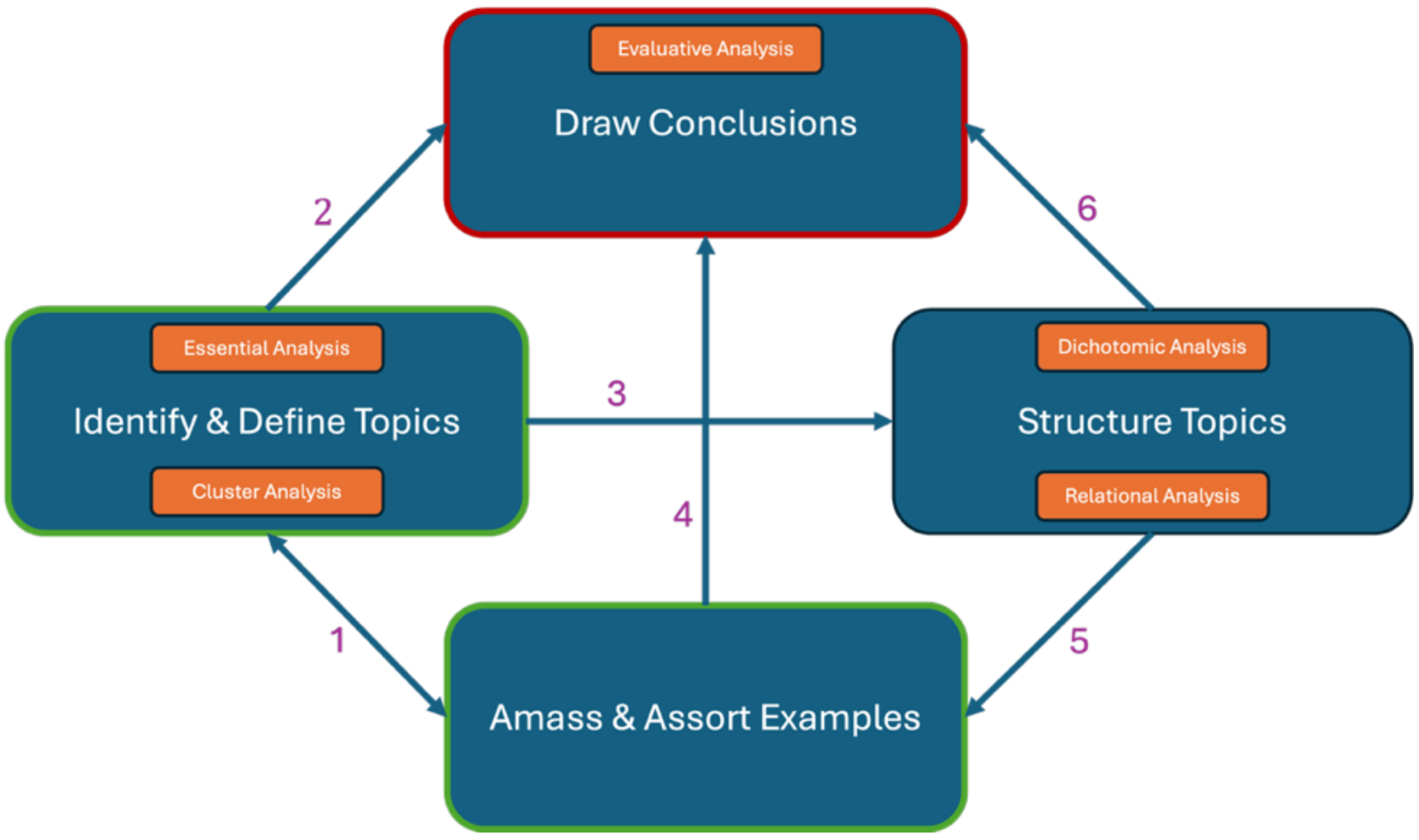

*Figure 1: Reflective analysis as a set of four interacting processes with embedded cores.*

*Identify & Define Topics*, Steps 2 and 5 mapped to *Amass &Assort Examples*, Step 4 mapped to *Structure Topics*, and Step 6 mapped to *Draw Conclusions*.

The newly identified cores of the three original forms of analysis are shown explicitly in this figure within the processes they support – essential analysis as part of Identify & Define Topics and both dichotomic and relational analysis as part of Structure Topics – while leaving open the possibility of other forms of analysis joining them. For example, evaluative analysis is now explicit within Draw Conclusions and *cluster analysis* (e.g., Michalski, 1980), a preexisting form of analysis that was previously buried within dichotomic analysis, is now explicit within Identify & Define Topics. As part of this rethinking, Step 3 from the reflective method – of identifying essences – is now expressed more generally in terms of defining topics, opening up the possibility of reflective analyses ultimately including other means of defining the topics of interest.

Processing in Figure 1 can start with either Identify & Define Topics or Amass & Assort Examples (both outlined in green). When applying essential analysis to preexisting topics – such as architectures, symbols, intelligence, and theories – processing begins with Identify & Define Topics. In contrast, when identifying core dichotomies, processing begins with Amass & Assort Examples. Yet, whichever of the two processes is the starting point, it is possible to move freely back-and-forth between them (via link $\overleftrightarrow{1}$) to enable cycles of identifying and defining topics mixed with amassing and assorting examples of them.

Structure Topics is then fed by Identify & Define Topics (via link $\overrightarrow{3}$). There is also a link from Structure Topics back to Amass & Assort Examples (via link $\overleftarrow{5}$), enabling Step 5 of the original reflective method plus more amassing and assorting of examples. Draw Conclusions is the end of the processing pipeline (outlined in red). It can be fed by any or all of the other three processes (via links $\overrightarrow{2}$, $\overrightarrow{4}$, and $\overrightarrow{6}$).

In this form, the four processes that make up reflective analysis can be applied flexibly to problems that would benefit from one or more of the embedded analytical cores. One new such application can be found in Section 7. However, the same is also true for the three original forms of analysis, each of which turns out to involve not only multiple processes but also multiple cores. The original form of essential analysis focused on the essential core, but it also leveraged the evaluative core (via links $\overleftarrow{1}$ and $\overrightarrow{2}$) for both intelligence and theories. For intelligence it further performed some simple structuring of the three levels of intelligence (via link $\overrightarrow{3}$) while drawing a few conclusions from this structuring (via link $\overrightarrow{6}$). The original form of dichotomic analysis combined the essential, dichotomic, and evaluative cores – plus cluster analysis – (via links $\overleftrightarrow{1}$-$\overrightarrow{6}$). The original form of relational analysis combined the essential, relational, and evaluative cores (via links $\overleftarrow{1}$, $\overrightarrow{3}$, and $\overrightarrow{6}$).[9] The development of reflective analysis thus clarifies the internal structure of these three previously monolithic forms of analysis while highlighting how their cores can be used flexibly in combination.

[9] Although the application of Draw Conclusions here includes a bit of the evaluative core, in considering the strength of methods used across disciplines, much more of it involved other forms of conclusions about disciplines and the relationships among them.

Given its composite nature it is hard to say anything substantive about the applicability of the reflective method. Perhaps the most that can be said is that it should be applicable to any problems for which some set of its cores is applicable. There is also, however, room to extend reflective analysis with other cores if/when they are needed, and thus to further extend its overall applicability.

**7. Analyzing a Pair of Dichotomies**

Two of the knottier dichotomies that made an appearance in the earlier work on dichotomic analysis were *procedural versus declarative* (e.g., Ryle, 1946; Winograd, 1975; Johnson-Laird, 1977; Woods, 1981; Squire 2004) and *implicit versus explicit* (e.g., Graf & Schacter, 1985; Sun, 2004; Squire & Dede, 2015; Sun, 2016). Each has a rich history on its own, but there is also an interesting interaction between them.

In the earlier work, these two dichotomies were introduced simply as examples of core dichotomies – the former for (a)symmetric, when considered as procedural versus declarative in the context of memory or learning or programming; and the latter for an alignment of all three, when considered as implicit versus explicit in the context of memory or processing. Here, the idea is to flip this on its head, foregrounding these two dichotomies as topics rather than backgrounding them as examples, and thus both to provide a fresh example of reflective analysis and to see what such an analysis can reveal about them. The results were foreshadowed in the earlier work on dichotomic analysis but in a more informal manner.

The general pathway taken by this analysis starts by identifying the topics, which is just the two dichotomies, and then moving to amass examples of each (via link $\overrightarrow{1}$). These examples are then used as the basis for defining each topic (via link $\overleftarrow{1}$), with some conclusions being drawn directly from these definitions (via link $\overrightarrow{2}$). The relationship between the two definitions is also structured through a different pathway (via link $\overrightarrow{3}$), leading to further conclusions (via link $\overrightarrow{6}$). Of the analytical cores, only essential analysis is used here. Yet, both Structure Topics and Draw Conclusions make appearances, just in forms that do not implicate their identified cores. For simplicity, the processing prior to Structure Topics will now be described separately for each of the two dichotomies before they are jointly structured.

For procedural versus declarative, moving to Amass & Assort Examples (via link $\overrightarrow{1}$) yields procedural versus declarative representations (knowing how versus what), procedural versus declarative memories (storing how versus what), procedural versus declarative learning (learning how versus what), procedural versus declarative programming (specifying how versus what), and procedural versus declarative semantics (defining meaning in terms of how versus what). Given these examples, returning to Identify & Define Topics (via link $\overleftarrow{1}$) yields the obvious essence for procedural versus declarative of *how versus what*. However, given that Rosenbloom (2026) already mapped the memory, learning, and programming examples onto the core dichotomy of (a)symmetric, this then becomes an alternative worth considering.

In favor of (a)symmetric as the essence is that it is more directly computational and thus also more directly mappable to varieties of technologies. Consider the dichotomy of heteroassociative versus autoassociative neural networks – a specialization of heteroassociative versus autoassociative memories – which distinguishes, for example, feedforward neural networks (Rumelhart *et al*., 1986a) from Boltzmann machines (Ackley *et al*., 1985). On its own it doesn't leap out as an example of how versus what. Yet it straightforwardly maps to (a)symmetric, and through this route can be seen to yield a basis for procedural versus declarative memory, with heteroassociative producing what comes next via feedforward processing and autoassociative filling in what content is missing.

If the essence of procedural versus declarative were just how versus what, then the definitional space might contain merely the set of procedural versus declarative dichotomies, covering at least representations, memories, learning, programming, and semantics. These yield a compact range of variations around the topic's essence. However, with the essence as (a)symmetric a much larger definitional space comes into focus. As mentioned in Section 3, a total of thirty-one preexisting dichotomies have so far been identified that map onto (a)symmetric, of which procedural versus declarative memory, learning, and programming were just three. The others included, for example, rules versus logic, reaction versus reflection, policies versus planning, arithmetic circuits (Darwiche, 2002) and sum-product networks (Poon & Domingos, 2011) versus probabilistic graphical models (Koller & Friedman, 2009), function-based reasoning versus model-based reasoning (Choi & Darwiche, 2018), heteroassociative versus autoassociative neural networks, skill versus knowledge, backpropagation (Rumelhart *et al*., 1986a) versus Hebbian learning (Hebb, 1949), classification versus clustering, discriminative versus generative learning (e.g., Jebara, 2004), primitive emotions versus cognitively elaborated emotions (Ortony *et al*., 2005), and causal intervention versus causal counterfactuals (Pearl & Mackenzie, 2018). Many of these are not obvious kin of procedural versus declarative – or even how versus what – at least on the surface, but the connection can be made through an essence of (a)symmetric.

For implicit versus explicit, moving to Amass & Assort Examples (via link $\vec{1}$) yields a more confusing picture, as over the years this dichotomy has been identified divergently as procedural versus declarative memory (e.g., Squire & Dede, 2015), automatic versus deliberate retrieval (e.g., Graf & Schacter, 1985), neural network versus symbolic processing (e.g., Sun, 2016),[10] and unconscious versus conscious memory (e.g., Squire & Dede, 2015). Moving back to Identify & Define topics (via link $\overleftarrow{1}$), the first three map, respectively, onto the core dichotomies of (a)symmetric,

[10] Sun (2016) also distinguishes procedural versus declarative but refers to this as action-centered versus non-action centered, which is then orthogonal to how implicit versus explicit is used.

(un)controlled, and (meta)data, suggesting that implicit versus explicit is an only vaguely understood dichotomy, much like (un)conscious.

Still, some kind of sense can be made of this diversity of notions by viewing implicit versus explicit as a particular alignment along these three core dichotomies; that is, explicit can be taken to require all three of symmetric, controlled, and data, whereas anything that is asymmetric or uncontrolled or metadata is then implicit.[11] These two formulations thus make sense as the essences for explicit and implicit, respectively, with the latter appropriately amounting to the logical negation of the former. No definitional space of interest has yet been identified here.

This essence further raises the question, in Draw Conclusions (via link $\vec{2}$), of just what within a cognitive system can actually achieve the particular combination of attributes required for explicit. As characterized in the Common Model of Cognition (Laird *et al.*, 2017), working memory is a short-term, symbolic, declarative memory that is the locus for omnidirectional communication among the other modules in the cognitive system. Most particularly, it is the only such memory that is directly accessible by procedural memory for driving controlled decision making. Working memory thus aligns well with the first three ways of thinking about the explicit side of the dichotomy. It also thus aligns well with access consciousness (Block, 2005), as characterized for example by Global Workspace Theory (Baars, 1988), thus bringing in the fourth way of thinking about the dichotomy – unconscious versus conscious memory – at least for access consciousness.[12] It says nothing, however, about phenomenal consciousness; that is, the more subjective side of conscious experience (Block, 2005).

Shifting now to Structure Topics (via link $\vec{3}$), it is first worth noting that a direct connection between the two dichotomies has already been made in the literature (e.g., Morsella & Poehlman, 2013; Squire & Dede, 2015). However, whereas this literature assumes a one-to-one relationship between the two dichotomies, the conclusion drawn here (via link $\vec{6}$) is that only the (a)symmetric aspect – associated with procedural versus declarative memory – is shared. Thus, although this commonality comprises all of the first topic, it only comprises 1/3 of the second.

Overall, this analysis depends on all four processes that make up reflective analysis, via links $\overleftrightarrow{1}$, $\overrightarrow{2}$, $\overrightarrow{3}$, and $\overrightarrow{6}$. In so doing it leverages the core of essential analysis, while structuring the topics in a way that depends neither on dichotomic analysis's formation of cross products nor relational analysis's combination of disciplines, and drawing important conclusions that do not depend on evaluative analysis's use of relative preferences. Does this indicate the implicit existence of other cores that should be made explicit in Structure Topics and Draw Conclusions, with the latter perhaps concerned with the overlap among topics? Possibly, although this one example on its own seems insufficient as a basis for developing such cores.

[11] This alignment is similar in spirit but quite distinct in detail from Bargh's (1994) breakdown of automaticity into the "four horsemen" of "Awareness, intention, efficiency, and control in social cognition."

[12] Rosenbloom *et al*. (2025) showed how the Common Model notion of working memory also aligns with key features of three other leading theories of consciousness: the global feedback signal of Recurrent Processing Theory (Lamme, 2006), the highest-level representations of Predictive Processing and Neurorepresentationalism (Hohwy & Seth, 2020; Pennartz, 2022), and the most integrated representations in Integrated Information Theory (Tononi *et al*., 2016).

## 8. Conclusion

This article has been concerned with how to move from scientific controversies to structured understandings of both the controversies themselves and the disciplines within which the controversies occur. Building on the earlier development of essential, dichotomic, and relational analysis, plus their synthesis into the reflective method, the focus here has been on a deeper understanding of these prior developments; a diagrammatic reformulation of the reflective method, as reflective analysis; and a reconceptualization of how the three initial forms of analysis relate to it, as analytical cores embedded within its four interlinked processes. Two additional analytical cores that were submerged within the original forms of analysis – evaluative analysis and cluster analysis – were also identified in the process, with the possibility remaining open of ultimately identifying even more cores, such as for structuring overlapping topics (Section 7).

In the process, the role of reflective analysis in leveraging scientific controversies for structured understanding has been identified, the nature of the definitions yielded by essential analysis has been clarified, as a form of neoclassical concept that is in service of metascientific utility rather than cognitive modeling or philosophy of mind; and the range of approaches used to yield these definitions, both inductive and deductive, has been made explicit. New examples of the application of both relational and reflective analysis have been included as well. More importantly, recasting the original three forms of analysis in terms of their cores has facilitated identifying applicability criteria for them that provide insight into where else they might be useful beyond the current focus on AI, cognitive science, and computing.

## References ▪ Kaynakça